# Fine-Tuning Large Language Models and Evaluating Retrieval Methods for Improved Question Answering on Building Codes

MOHAMMAD AQIB[1], Mohd Hamza[2], Qipei Mei[1], Ying Hei Chui[1]

*Department of Civil & Environmental Engineering, University of Alberta, Edmonton, Alberta, Canada*[1]

*Aligarh Muslim University, Aligarh, Uttar Pradesh, India*[2]

**Abstract:** Building codes are regulations that establish standards for the design, construction, and safety of buildings to ensure structural integrity, fire protection, and accessibility. They are often extensive, complex, and subject to frequent updates, making manual querying challenging and time-consuming. Key difficulties include navigating large volumes of text, interpreting technical language, and identifying relevant clauses across different sections. A potential solution is to build a Question-Answering (QA) system that answers user queries based on building codes. Among the various methods for building a QA system, Retrieval-Augmented Generation (RAG) stands out in performance. RAG consists of two components: a retriever and a language model. This study focuses on identifying a suitable retriever method for building codes and optimizing the generational capability of the language model using fine-tuning techniques. We conducted a detailed evaluation of various retrieval methods by performing the retrieval on the National Building Code of Canada (NBCC) and explored the impact of domain-specific fine-tuning on several language models using the dataset derived from NBCC. Our analysis included a comparative assessment of different retrievers and the performance of both pre-trained and fine-tuned models to determine the efficacy and domain-specific adaptation of language models using fine-tuning on the NBCC dataset. Experimental results showed that Elasticsearch proved to be the most robust retriever among all. The findings also indicate that fine-tuning language models on an NBCC-specific dataset can enhance their ability to generate contextually relevant responses. When combined with context retrieved by a powerful retriever like Elasticsearch, this improvement in LLM performance can optimize the RAG system, enabling it to better navigate the complexities of the NBCC.

## 1. Introduction

The construction industry plays a vital role in an economy as it significantly influences capital flow and creates employment opportunities [1]. They are also indirectly linked with other economies; hence, an increase in the construction sector enhances the demand for other sectors [2]. Execution of construction projects is always challenging, and their processes are influenced by a certain set of rules and regulations, which are mentioned in building codes. Building codes are legal documents developed by government agencies to control building practices through specific technical and functional criteria [3]. Given the extensive content encompassing a wide range of technical standards and regulations, querying them efficiently can be a challenging task for professionals such as architects, engineers, and contractors. The need for accurate and timely access to relevant sections of the building codes is critical, particularly when dealing with complex building projects where specific code requirements must be identified quickly. However, the sheer volume and technical nature of such complex documents present challenges. It has been observed that compliance checking over vast building codes is one of the major bottlenecks in the design and construction process [4]. The traditional way of querying building codes is an empirical-based or keyword matching-based search, and it may be time-consuming and produce inaccurate results. Users have to go through all the results and identify the correct answer from the pool of answers. Moreover, users must be aware of the exact word while using keyword-based searching [5]. These issues highlight the necessity for more advanced systems that can interpret and retrieve information based on the context and meaning of the queries, rather than relying solely on surface-level text matching. Such systems would be capable of answering questions when asked in natural language and would also eliminate the dependency on exact words to get an answer. These systems should find the answers in complex building codes and reduce efforts, time, and errors. The ability of such systems to understand queries and the context of the building codes would significantly improve the user's productivity, providing quick and precise access to critical regulatory information. One promising solution to the issue is to build a Question-Answering system by utilizing the power of Large Language Models (LLMs) in the Retrieval-Augmented Generation (RAG) framework. RAG utilises powerful retrievers and text generational capabilities of LLMs to provide responses to the asked queries.

LLMs represent a significant improvement in the field of artificial intelligence and have transformed several fields [6]. LLMs are trained on a vast number of datasets, including books, articles, Wikipedia, and more, allowing them to learn a wide scope of knowledge [7]. They gained worldwide popularity after the release of an application by OpenAI, i.e., ChatGPT [8]. Recently, LLMs have shown potential usage in programming [9], legal [10], medical [11], and many other fields [12]. LLMs based on transformers such as GPT-4 [13], BERT [14], PaLM [15], and LLaMA [16] have revolutionized NLP by providing impressive and human-comparable performance over various benchmarks [17]. Additionally, LLMs are used for content generation and automation, allowing businesses to produce high-quality articles, product descriptions, and marketing materials quickly and consistently [18].

LLMs are transforming the construction industry by enhancing efficiency, accuracy, and decision-making across various aspects of project management. These data-driven models are revolutionizing traditional practices and providing innovative solutions to longstanding challenges. Notable applications include safety hazard identification in construction using vision language models [19], compliance checks for regulatory documents [20], safety assessments using Visual Language Models (VLMs) [21], ergonomic assessment of construction workers using VLMs [22], and contract risk identification [23]. Additionally, LLM-based frameworks have enabled real-time issue identification and mitigation in construction meetings [24], while automated Building Information Modeling (BIM) compliance checks through LLMs further highlight their growing role in the industry [25]. Another example of utilizing LLMs in civil engineering is their use in developing frameworks and implementations for answering questions related to building codes and standards [26]. With rapid advancements in AI technologies, LLMs and VLMs are poised to become integral to the future of the construction sector, driving greater automation and efficiency.

Although LLMs have been very popular lately, domain-specific tasks are still an issue for them [27]. The major problem is that models generate wrong answers when they are queried outside of their training data on occasions, a condition called hallucination [28]. RAG can withstand such drastic issues caused by hallucination and is widely utilized to enhance the quality and efficiency of generated content nowadays [29]. RAG makes use of the retrieval methods and generation capabilities of LLMs to produce contextually full and relevant responses to the questions asked from external knowledge databases [30]. In a RAG framework, retrievers play a critical role by extracting the relevant context from a large document in response to a query. This retrieved context is then passed to the LLM, which generates the final answer. The effectiveness of a RAG system relies heavily on the quality of the retrieval process.

Fine-tuning is another technique for adapting pre-trained LLMs to new tasks and reducing hallucinations by continuing training on new data [31]. Fine-tuning involves training pre-trained models on task-specific datasets, which helps in making the model capable enough to perform better on the desired task. It has been found that fine-tuning the LLM in an RAG system can help improve its performance, as the LLM is responsible for text generation utilizing the retrieved information [32]. However, full fine-tuning, which initializes the model with pre-trained weights and updates all of them, becomes impractical when dealing with models having billions of parameters [33]. Great efforts are put into avoiding training all parameters of the model and promoting an efficient fine-tuning process [34]. There are several techniques that are utilized to fine-tune models efficiently and can resolve the issue of heavy computational dependency that arises due to full fine-tuning. One of the most popular techniques for fine-tuning language models is parameter-efficient fine-tuning (PEFT), which efficiently adapts language models to specific downstream tasks while minimizing the constraints of dependency on computational resources and storage requirements. There are several methods of PEFT which basically freeze most of the parameters in a model and modify a few parameters, and they include BitFit [35], Visual Prompt Tuning [36], Low Rank Adaptation (LoRA) [37], and Adapter [38]. Even though PEFT modifies only a few parameters while training, it has been found that post fine-tuning using PEFT gives comparable performance to full-fledged fine-tuning [39].

Considering the crucial role of RAG in developing efficient Question Answering systems for building codes, this research focuses on integrating the fine-tuning technique and evaluating the key components of RAG, which are essential for the system's performance. Specifically, the paper identifies the best retriever and also explores how fine-tuning can significantly improve the generative capabilities of LLMs. To perform experiments on components of RAG, we have utilized the dataset derived from the NBCC. The NBCC serves as a comprehensive set of guidelines that regulates the design, construction, renovation, and demolition of buildings [40].

The paper is structured as follows. In Section 2, the process begins with defining the RAG and its workflow, detailing how RAG utilizes retrievers and LLMs to address users' queries. It also includes detailed information on different retrievers that were used, utilized fine-tuning techniques, and evaluation metrics used for testing retrievers and language models. Section 3 delves into dataset pre-processing, preparation, and statistical insights of the prepared dataset. Moving forward, Section 4 discusses the evaluation results of retrievers, pre-trained models, and fine-tuned models. Section 5 discusses the key findings and conclusions of the paper.

## 2. Methodology

*2.1 Retrieval Augmented Generation (RAG)*

RAG is a hybrid framework that harnesses the power of retrievers and LLMs to generate responses against a query from a database [41]. This approach first identifies relevant documents to query from a large database and then generates the final response to the query asked using LLMs and retrieved information. The retriever's ability to extract relevant information and the LLM's generative capabilities to generate coherent and contextually accurate responses greatly influence the performance of RAG. RAG can be beneficial for cases that require a regular point of reference over complex documents. Building code is one of the examples of a complex document frequently queried by engineers, and RAG can save them significant time and effort typically spent navigating and interpreting it. Usually, a RAG system breaks down the document into chunks, which are then converted and stored as embeddings. Embeddings are vector representations of data that capture their semantic meaning. These embeddings are utilized by the RAG system to perform efficient and meaningful search against the query asked. When a query is received, it is encoded into embeddings and compared with the stored embeddings of the document to retrieve the most relevant information related to the query. Comparison and finding relevant information from the document are conducted by retrievers utilizing different techniques. This relevant information and the query are finally passed to the LLM for final answer generation. The detailed workflow of RAG for a building code, depicting different components, is illustrated in Figure 1.

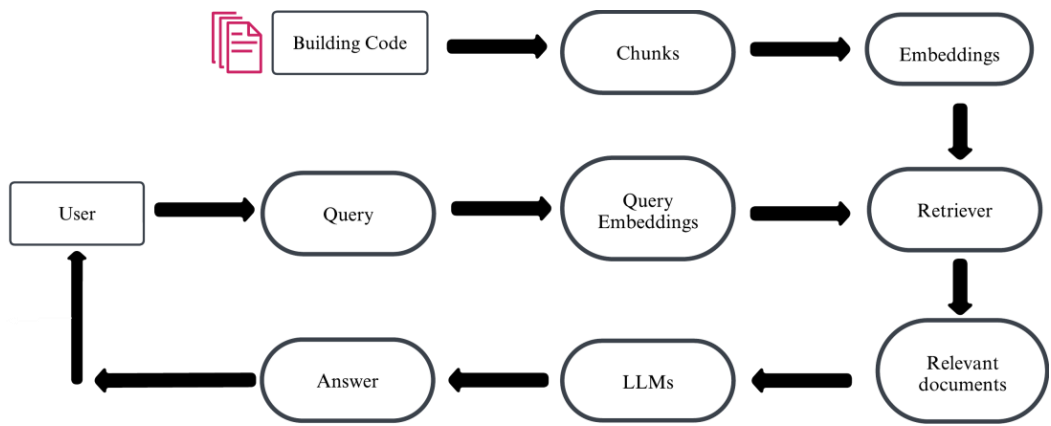

**Figure 1.** Workflow of RAG

*2.2 Retrieval Techniques*

In a RAG system, retrievers are responsible for retrieving relevant documents from the databases against a query. The correctness of the generated answer depends on the retrieved context, and therefore, it is very crucial to find out the best retrievers among all available. There exist several retrieval algorithms that can be used in a RAG system, sparse and term-based methods including Best Match 25 (BM25) [42], Term-Frequency Inverse Document-Frequency (TF-IDF) [43], dense, vector-based approaches like Dense Passage Retrieval (DPR) [44], etc. Sparse and term-based methods are based on the exact occurrence and frequency of terms within documents [45], while dense and vector-based methods use deep learning techniques to project text into dense vector spaces. The sparse methods are mostly faster and more interpretable; thus, they become suitable in scenarios where exact term matching is the key. Dense and vector-based methods capture semantic similarities between queries and documents, even when they do not share exact keywords [44]. Though dense methods are much more computationally expensive, they are much better at modeling deeper semantic relationships and tasks that require a nuanced understanding of language [44].

In this paper, we have assessed a range of retrieval techniques, including Term Frequency - Inverse Document Frequency (TF-IDF) and Best Match-25 (BM25) for conventional text retrieval, Elasticsearch (ES) for comprehensive text search, Dense Passage Retrieval (DPR) paired with Facebook AI Similarity Search (FAISS) for efficient similarity search, S-BERT combined with cosine similarity for semantic matching, and DPR with Annoy Index for approximate nearest neighbor search. In this paper, we have used 1436 triplets of Context-Question-Answer (CQA) datasets derived out of NBCC to find the best retrievers for our case as described in Section 3. Queries were passed to each retriever technique and the best possible top-1, top-3, top-5 and top-10 documents were retrieved from the NBCC document. Top-1, top-3, top-5 and top-10 retrieval settings refer to the number of most relevant results returned by the retrieval system in response to a query. Top-1, top-3, top-5 and top-10 returns the single best, the best 3, the best 5 and the best 10 relevant documents out of NBCC to a query respectively. Detailed explanations of each retrieval system are provided in subsequent subsections. To evaluate the performance of the retrieval algorithms, each retriever was tasked with identifying the top-1, top-3, top-5 and top-10 most relevant documents out of NBCC, in response to a given question from the derived Context-Question dataset. The relevance of the retrieved documents was assessed by comparing them with the reference contexts using BERT scores. BERT metric includes BERT Precision, BERT Recall, and BERT F1-score, allowed for a detailed analysis of how well the documents retrieved from

NBCC aligned with the true context, providing insights into the efficacy of each retriever in accurately capturing the intended information.

*2.2.1 Term Frequency-Inverse Document Frequency (TF-IDF)*

TF-IDF is a classical technique for retrieval. It models the relevance of a document to a query by computing term frequency and inverse document frequency. While TF measures the frequency of terms in a document, IDF measures the importance of terms in the entire corpus. Multiplying TF and IDF gives a score of how important the term is to the document regarding the corpus. The higher the score, the more relevant to the query.

Here, the document is treated as a collection of chunks, each chunk representing a distinct retrieval unit. The TF measures how often a query term appears within a particular chunk, normalized by the total number of terms in that chunk. IDF is mainly involved in identifying the importance of the relevancy of terms in the entire corpus. For a query Q and a chunk C of documents, the relevance score is computed as:

$$TF-IDF(Q,C) = \sum_{t \in Q \cap C} TF(t,C) \cdot IDF(t) \tag{1}$$

$$TF(t,C) = \frac{f_t^C}{T_C} \tag{2}$$

$$IDF(t) = \log\left(\frac{N}{n_t + 1}\right) \tag{3}$$

where $f_t^C$ is frequency of term t in chunk $C$, $T_C$ is total number of terms in $C$, $N$ is total number of chunks in the document and $n_t$ is number of chunks containing term t.

*2.2.2 Best Match-25 (BM-25)*

The BM25 is a probabilistic-based retrieval model and an extension of the TF-IDF model. It computes the relevance of documents by measuring the query term frequency in a document, and that term's distribution across the entire corpus. It creates a score by considering TF, IDF, document length and query terms weight. BM25 creates a balanced measure of term importance by adjusting for document length using parameters controlling term saturation and document length normalization. This makes BM25 pretty good at ranking documents based on a query. For a query Q and a chunk C, the relevance score is calculated by:

$$\text{BM25}(Q, C) = \sum_{t \in Q \cap C} \text{IDF}(t) \cdot \frac{\text{TF}(t, C) \cdot (k_1 + 1)}{\text{TF}(t, C) + k_1 \cdot \left(1 - b + b \cdot \frac{|C|}{\text{avg}[C\ |]}\right)} \quad (4)$$

where *t* represents each term shared between the query and the chunk, *IDF(t)* measures the term's importance, *TF(t,C)* quantifies frequency of t in chunk C, $k_1$ is a tuning parameter (between 1.2-2), *b* adjusts for length normalisation and is set to 0.75, and $|C|$ represents number of words in chunk and avg $|C|$ in average length of chunk across the corpus.

*2.2.3 Elasticsearch*

Elasticsearch is a compelling, Apache Lucene-based search platform. It is widely used for indexing and retrieving documents from large datasets due to its scalability and speed. In Elasticsearch, this is achieved through inverted indexing for efficient document retrieval that contains query terms. Unlike traditional sequential scanning, the inverted index organizes data in a way that facilitates fast query processing, even over large and complex corpora such as building codes. Elasticsearch begins with pre-processing techniques which involve tokenization, lowercasing, stemming and lemmatization. Tokenization is the process where text is split into its constituent words or phrases, and lowercasing involves normalizing text to ensure case insensitivity. On the other hand, stemming and lemmatization involve the conversion of words into their respective root or base forms. Now, a pre-processed document is converted into a searchable inverted index structure, which enables efficient retrieval while supporting advanced features such as phrase matching and term proximity. The inverted index is a data structure that maps terms to the documents containing them, enabling rapid lookup and retrieval. Moreover, it runs relevance BM25 scoring algorithms to rank documents by their relevance order to a query. A detailed overview of the functionality of Elasticsearch is mentioned in Figure 2.

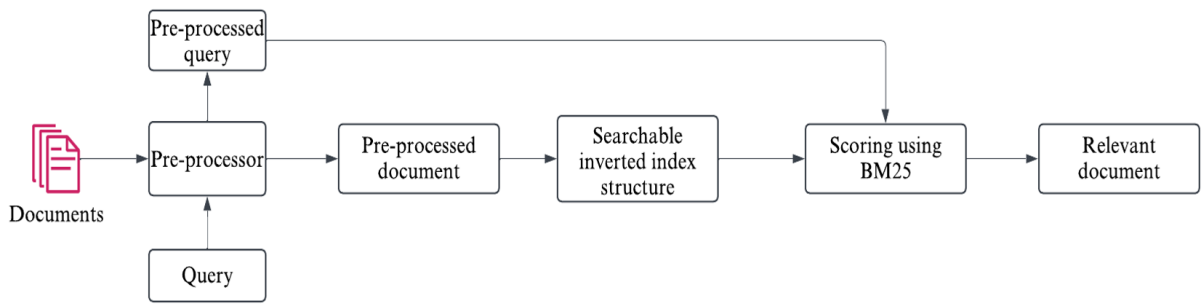

**Figure 2.** Workflow of Elasticsearch

*2.2.4 DPR with Facebook AI Similarity Search (DPR-FAISS)*

Dense Passage Retrieval (DPR) is a neural-based retrieval model that learns dense vector representations of the provided text. It encodes the passages and queries into high-dimensional vectors with the help of pre-trained transformer models. FAISS is a library for efficient similarity search and clustering of dense vectors [46]. It provides a fast approximate nearest neighbor search using vector representations via indexing and querying. Putting together, DPR and FAISS enable precise and scalable retrieval of relevant documents powered by semantic similarity.

The document is split into smaller passages to ensure compatibility with the model input limits. Each passage is then tokenized and then encoded into a dense vector using the DPR model. For fast and scalable similarity searches, FAISS stores these embeddings of passage into an efficient data structure. This allows it to quickly compare the query embeddings with the passage embeddings and retrieve the most similar one. The formula used for determining relevant passages for a query is mentioned Eq. 5:

$$\text{Similarity} = \vec{q} \cdot \vec{p} \tag{5}$$

where $\vec{q}$ represents the query embeddings, and $\vec{p}$ represents the passage embeddings. Higher similarity scores indicate stronger alignment between the query and passage.

*2.2.5 Sentence-BERT with Cosine Similarity (S-BERT-Cos)*

Sentence-BERT is another variant of the BERT model that is specifically optimized to produce sentence embeddings [47]. It creates dense vector representations of sentences, capturing the semantic meaning in a sentence. These semantic similarities are computed further on the vectorized queries and documents using cosine similarity. By calculating the cosine of the angle between vectors, S-BERT with cosine similarity efficiently ranks documents based on their semantic relevance to the query.

The retrieval process begins with indexing, where the document is segmented into smaller chunks. These small and manageable chunks are converted into dense vector embeddings using the S-BERT model. Generated embeddings capture semantic relationships within text. When a query is passed, it is first converted to dense vector embeddings using the same S-BERT model. The similarity between the embeddings of the query and each passage is calculated using cosine similarity. The retrieval framework employs a top-k selection mechanism, where the cosine similarity scores are used to rank all passages, and the top-k most similar passages are selected as relevant. Mathematically, cosine similarity is defined as:

$$\text{Cosine Similarity } = \frac{\vec{q} \cdot \vec{p}}{\|\vec{q}\|\|\vec{p}\|} \tag{6}$$

where $\vec{q}$ represents the query embedding, and $\vec{p}$ represents the passage embedding. The numerator denotes the dot product of the two vectors, while the denominator normalizes the values using their respective magnitudes.

*2.2.6 DPR with ANNOY Index (DPR-AI)*

Similar to the approach of DPR-FAISS, DPR encodes documents and queries into dense vectors with the help of a pre-trained transformer model. The Approximate Nearest Neighbour Oh Yeah (ANNOY) library was created by Erik Bernhardsson and is meant for searching points in a space that are close to a given query point [48]. It is an indexing library aimed at fast approximate nearest neighbor searches, especially in very high-dimensional vector spaces, which is achieved by constructing trees for efficient querying. ANNOY builds multiple binary trees, where each tree is constructed by repeatedly splitting the data points along random hyperplanes. These hyperplanes are chosen to maximize the separation between data points, effectively partitioning the vector space into regions. During a query, ANNOY traverses these trees, narrowing down the search space to a subset of relevant regions. With 10 trees in current setup, ANNOY finds the items most similar to the query asked and doesn't search through all the data points. Instead, it looks through the smaller groups created by the trees. By checking only a few of these groups, Annoy narrows down the search to find the nearest neighbors quickly. This significantly reduces the computational complexity compared to exhaustive searches. The method leverages a dot-product similarity metric to measure the relevance between the query and indexed passages. Thus, DPR powered with ANNOY allows fast retrieval of relevant documents by approximating the nearest neighbors within this vector space, balancing the speed and accuracy of the search.

*2.3 Fine-tuning Large Language Models*

Fine-tuning LLMs is essential for adapting pre-trained models to specific downstream tasks. Among various fine-tuning strategies, PEFT methods such as LoRA offer significant advantages. These methods minimize computational overhead and memory usage by updating only a small number of parameters instead of all weights present in the pre-trained model. Moreover, they cut down the dependency on large dataset required in full-fledged fine-tuning. In particular, LoRA introduces trainable low-rank matrices, allowing the model to learn task-specific patterns efficiently while preserving the original weights. LoRA trains only the weights of the newly added low-rank matrices while keeping the weights of pre-trained model frozen. This approach preserves the original knowledge of the model and therefore significantly reduces the risk of catastrophic forgetting. Concretely, instead of updating the original weight matrix W, LoRA approximates the update as a product of two smaller trainable matrices A and B, such that:

$$Output = W \cdot x + A \cdot (B \cdot x) \tag{7}$$

Here, $x$ is the input vector, $W$ is the frozen pre-trained weight matrix, A and B are trainable low rank matrices introduced by LoRA. A and B are relatively very small matrices as compared to W, which is why LoRA is memory efficient.

In our study, LoRA was applied to the core components of the Transformer architecture present in LLMs. Transformers were first introduced by Vaswani et al. [49] and are considered to be backbone of LLMs. The transformer model employs stacked layers of self-attention and feedforward operations to capture dependencies within input sequences. Key elements such as query, key, and value projections form the foundation of the attention mechanism, enabling the model to focus on relevant parts of the input. These mechanisms are complemented by layer normalization, gating, and feedforward sub-layers to further enhance representation learning.

Figure 3 illustrates the encoder-decoder structure of the Transformer. Within each encoder layer, the model applies a self-attention mechanism where the input embeddings are first projected into query (q_proj), key (k_proj), and value (v_proj) vectors. These projections allow the model to compute attention scores, enabling it to identify and weigh the relevance of different tokens in the sequence. The results are then combined and passed through an output projection (o_proj) to form the contextualized representation. Following this, the information flows through a feedforward network that includes gating layers (gate_proj) and projection layers (up_proj and down_proj) to further refine the representations and introduce non-linearity. The decoder operates similarly but incorporates an additional attention mechanism that allows it to attend not only to previous outputs but also to the encoder's outputs. Like the encoder, the decoder also uses query, key, and value projections, along with feedforward and gating layers, to generate predictions step by step.

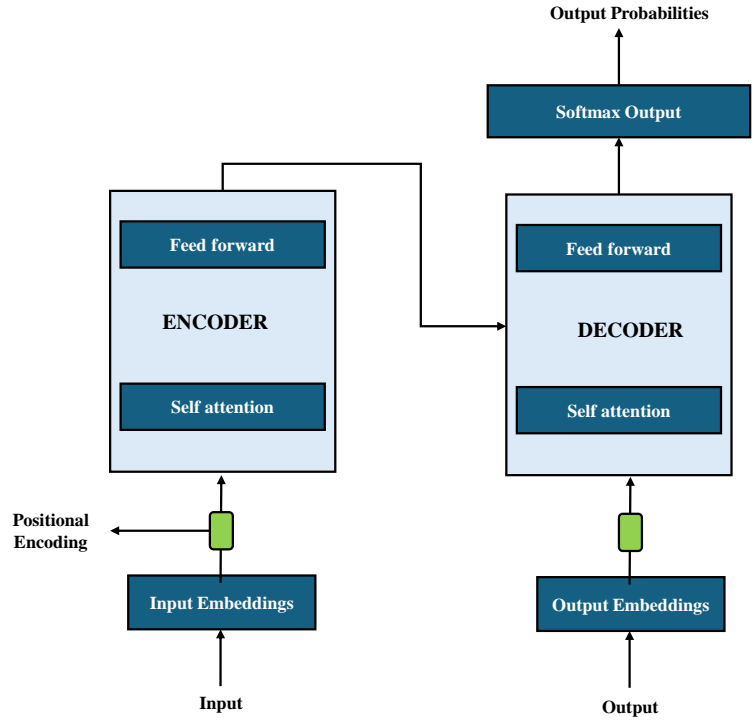

**Figure 3.** Simplified transformer architecture

Fine-tuning using LoRA and focussing on these key components, specifically the query, key, value, output, gating, and feedforward projections, we achieved targeted model adaptation on the dataset derived from NBCC. This approach enabled the model to efficiently adjust to task-specific features while minimizing the number of trainable parameters and reducing computational resource usage. The models evaluated and their respective trainable parameters are summarized in Table 1.

**Table 1.** Total and trainable parameters of model used.

| Model Name | Total Parameters (Billion) | Trainable parameter (Million) | Percentage Trained (%) |
|---|---|---|---|
| Qwen-2.5-7b-Instruct | 7 | 40.37 | 0.58 |
| Llama-2-7b | 7 | 93.98 | 0.57 |

| Model | | | |
|---|---|---|---|
| Llama 3.1-8b | 8 | 41.94 | 0.52 |
| Llama 3.2-1b-Instruct | 1 | 11.27 | 1.13 |
| Phi-3-Mini-4k-Instruct | 4 | 29.88 | 0.75 |
| Mistral-Small-24b-Instruct-2501 | 24 | 92.4 | 0.39 |

The fine-tuning process was conducted uniformly across all selected models, using a consistent set-up to ensure proper comparability of results. Each model was fine-tuned for 10 epochs. A detailed overview of fine-tuning set-up is described below:

- **Model Selection and Quantization:** Models to be fine-tuned were selected and configured to perform 4-bit quantization. Quantization techniques help in reducing the costs associated with LLMs by reducing computational and resource demands [50]. As a result, the models can process datasets and long input sequences more effectively. Moreover, maximum sequence length of 2048 tokens were used to accommodate extensive context while maintaining training stability.

- **Fine-Tuning Setup:** The fine-tuning process began by loading a pre-trained language model and injecting trainable LoRA matrices into select key component of attention mechanism and other transformer component such as q_proj, k_proj, v_proj, o_proj, gate_proj and others. These lightweight trainable LoRA matrices allowed for efficient training while keeping the original pre-trained model weights frozen, significantly reducing memory and computational requirements. To prepare the model for fine-tuning, gradient checkpointing was enabled, enabling support for longer sequences with reduced memory usage.

   The dataset was then formatted using the Alpaca-style prompt template, organizing the input as a combination of context, question, and answer. This formatted data was tokenized and mapped into the appropriate structure for fine-tuning. Fine-tuning was performed using the SFTTrainer from the trl library, which specifically updates only the LoRA parameters while leaving the rest of the model untouched. An overview of fine-tuning process is shown in figure 4.

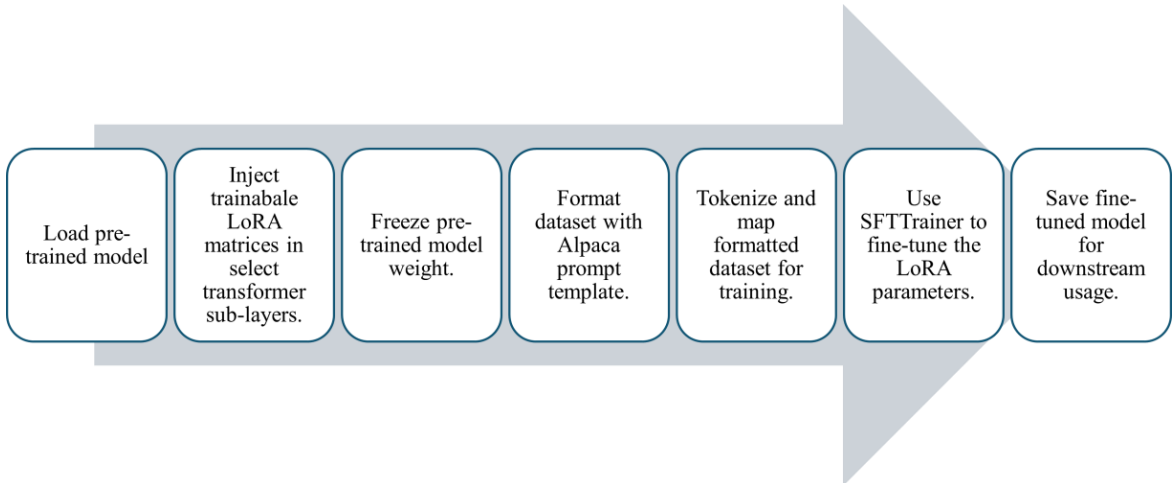

**Figure 4.** Workflow of fine-tuning process using LoRA.

The training process was configured for a total of 10 epochs, allowing the model to make multiple passes over the dataset to reinforce learning and improve generalization. A batch size of 2 was chosen to accommodate the memory limitations typically encountered with quantized large language models, and gradient accumulation steps were set to 4 to simulate a larger effective batch size and stabilize training. A learning rate of $2\times10^{-4}$ provided a balanced pace for convergence, and the AdamW optimizer with 8-bit precision was used to reduce memory usage while maintaining adaptive learning capabilities. Additionally, a linear learning rate scheduler was applied to gradually decrease the learning rate, aiding in smoother convergence. Training progress was logged every 100 iterations to monitor performance and stability throughout the process.

- **Computational Resources and Libraries:** All experiments were conducted on a 64-bit machine equipped with an 80-core Intel Xeon Gold 6242R processor running at 3.1 GHz, with 128 GB of RAM, and operating on Linux Ubuntu 22.04. Locally run LLMs were inferenced using 4-bit quantization on NVIDIA A6000 GPU with 48 GB of VRAM. The experimental setup was implemented using Python 3.10 and utilized libraries such as PyTorch, HuggingFace Transformers, Datasets, PEFT and TRL.

Fine tuning was performed to align LLMs with the intricacies of content present in the NBCC and improve their text generational capabilities. The LLMs chosen for fine-tuning include Qwen-2-5, Llama-2, Llama-3.1, Llama-3.2, Phi-3-Mini and Mistral. Llama-2 [51] and Llama-3 [52] are LLMs released by Meta AI and come in different variants. Similarly, Qwen, Mistral, and Phi models were released by Alibaba [53], Mistral AI [54], and Microsoft [55], respectively. For fine-tuning and evaluation of these models, we have utilized training and testing datasets, respectively. The training dataset comprises CQA triplets from the NBCC, while the test dataset includes only Context-Question pairs, and the models are tasked with generating the corresponding answers. Both pre-trained and fine-tuned models were evaluated on the test dataset to generate answers for each Context-Question pair. Each set of generated answers was compared against the ground truth answers using the evaluation metrics described in Section 2.4. The calculated metric score was then used to compare the performance of pre-trained and fine-tuned models.

*2.4 Evaluation Metrics*

Evaluating the performance of text generation models and retrievers requires comparing predicted outputs to reference or ground truth sentences. The process typically begins with tokenization, where both the predicted and reference sentences are split into smaller units, such as words or sub-word units. This segmentation forms the basis for further comparisons, where individual tokens, or unigrams, are matched between the two texts. In some cases, higher-order n-grams (sequences of tokens) are also considered to account for context and more complex patterns. To improve the robustness of the evaluation, additional techniques like stemming (which reduces words to their root forms), and synonym matching are employed to address variations in word forms or meaning. Another critical aspect of modern evaluation involves embeddings, which are dense vector representations of words or entire sentences. These embeddings capture semantic meaning, helping to assess the similarity between words or sentences beyond their surface-level structure. Advanced metrics also leverage these embeddings to calculate similarity scores and to handle more nuanced comparisons.

Precision, recall, and F1 score form the foundation of the evaluation metrics employed. Precision measures the proportion of correctly predicted elements among all predicted elements, highlighting the model's accuracy in avoiding irrelevant outputs. Recall evaluates the proportion of relevant elements correctly identified by the model, ensuring comprehensive coverage of expected information. F1 score, the harmonic mean of precision and recall, balances these two metrics to provide a unified and robust performance measure. These discussed foundational concepts form the basis for several evaluation metrics.

The choice of metrics was guided by the need to evaluate both the semantic accuracy and completeness of the generated answers. BERT-based metrics (F1, Precision, and Recall) are crucial as they assess the model's ability to generate answers that not only match the reference at a surface level but also align with its intended meaning. Sentence Mover's Similarity was included to capture the overall coherence between the generated and reference answers, while traditional metrics such as ROUGE, BLEU, and METEOR were used to provide additional insights into content overlap and linguistic variation. This combination of metrics ensures a comprehensive evaluation of the generated answers, focusing both on meaning and relevance, which are crucial for tasks involving context-based question answering. Detailed descriptions of each metric utilized are provided below.

*2.4.1 F1 score*

The F1 Score combines precision and recall into a single metric, making it a more balanced evaluation [56]. The F1 score for each pair of predicted and reference answers is computed based on the number of overlapping tokens. Let $P_i$ represent the predicted sentence and $R_i$ the corresponding reference sentence, where each answer is split into tokens. The F1 score is calculated as:

$$F1(P_i, R_i) = 2 \cdot \frac{\text{Precision}(P_i, R_i) \cdot \text{Recall}(P_i, R_i)}{\text{Precision}(P_i, R_i) + \text{Recall}(P_i, R_i)} \tag{8}$$

$$Precision(P_i, R_i) = \frac{|P_i \cap R_i|}{|P_i|} \tag{9}$$

$$\text{Recall}(P_i, R_i) = \frac{|P_i \cap R_i|}{|R_i|} \tag{10}$$

For the entire dataset of n predictions, the average F1 score is computed as:

$$\text{Average F1 Score} = \frac{1}{n} \sum_{i=1}^{n} F1(P_i, R_i) \tag{11}$$

where $|P_i \cap R_i|$ is intersection of unigrams present in both $P_i$ and $R_i$, $|R_i|$ represents total count of unigram in reference sentence and $P_i$ represents total count of unigrams in predicted sentence.

*2.4.2 BLEU Score*

BLEU measures the precision of n-grams in the predicted answer compared to the reference [57]. To compute BLEU in this study, we used the NLTK library's sentence_bleu function, considering unigrams and bigrams with equal weights (0.5 each). Moreover, to address cases where overly short predictions might achieve high n-gram precision, BLEU employs a brevity penalty (BP). The BP adjusts the BLEU score downward for predictions that are significantly shorter than the reference, ensuring that the score reflects both n-gram accuracy and the length adequacy of the generated answer.

$$BLEU(P,R) = BP(P,R) \cdot \exp\left(\sum_{n=1}^{N} w_n \cdot \log p_n\right) \quad (12)$$

$$BP(P,R) = \min\left(1, \frac{|P|}{|R|}\right) \quad (13)$$

where $p_n$ is the precision of n-grams of order n for n=1,2,..,N where N typically ranges from 1 to 4. In our case N=2 as we have considered only unigram and bigram. $w_n$ is the weight associated with each n-gram precision, which is typically equal to 1/N for equal weighting. For our case $w_n = 0.5$. Moreover, |P| and |R| are total tokens present in the predicted and reference sentences, respectively.

### 2.4.3 ROUGE-1 Score

The ROUGE (Recall-Oriented Understudy for Gisting Evaluation) Score measures the overlap between predicted and reference answers at the n-gram level [58]. Unlike BLEU, which focuses on precision, ROUGE primarily evaluates recall. It is computed for different n-grams, such as unigrams, bigrams, and longest common subsequences. We have utilised unigram for calculation of ROUGE score. Firstly, both the predicted and reference answers are tokenized into unigrams and then overlap is calculated for each word in reference answer. Mathematically representation of the ROUGE-1 is mentioned below. Numerator captures the overlap of unigrams between reference and predicted sentence while denominator denoted total number of unigrams in the reference summary.

$$\text{ROUGE-1} = \frac{\sum_{w \in R} \min.(Count_R(w), Count_P(w))}{\sum_{w \in R} Count_R(w)} \quad (14)$$

where $w$ represents a unigram (single word) in the reference sentence R, $Count_R(w)$ is the number of times the unigram w appears in R and $Count_P(w)$ is the number of times the unigram $w$ appears in the predicted sentence $P$.

### 2.4.4 Sentence Mover's Similarity (SMS)

It is text similarity metrics which advances over Word Mover's Similarity [59]. It measures how difficult it is to transform from one sentence to another by moving the words while considering semantic relationship between them. Therefore, lower score means the predicted sentence is more similar to reference or ground truth sentence. It uses word embeddings to capture semantic relationship between sentences. SMS is based on Earth Mover's Distance (EMD) which is a concept from optimal transport theory. EMD calculates minimum efforts needed to transport from one probability distribution to another. SMS is computed as the minimum cost to transport words from predicted and reference sentence using their embeddings. Mathematically it can be defined as:

$$\text{SMS}(S_{\text{pred}}, S_{\text{ref}}) = \min_{T \geq 0} \sum_{i,j} T_{ij} d(w_i, w_j) \tag{15}$$

where $S_{\text{pred}}$ is predicted sentence, $S_{\text{ref}}$ is reference sentence, $d(w_i, w_j)$ is cosine distance between the embeddings of word $w_i$ from $S_{\text{pred}}$ and word $w_j$ from $S_{\text{ref}}$. $T_{ij}$ is optimal transport matrix which determines how much of word $w_i$ should be transformed to match $w_j$.

### 2.4.5 METEOR Score

The METEOR (Metric for Evaluation of Translation with Explicit ORdering) Score considers synonyms, stemming, and word order in addition to n-gram matches. METEOR combines precision and recall by giving higher weights to exact matches and allowing for partial matches of synonyms or stemming variants [60]. It can match words with the same meaning and common root. It utilises penalty criteria which penalises the score based on word order.

$$\text{METEOR} = F_{\text{mean}} \cdot (1 - \text{Penalty}) \tag{16}$$

$$F_{\text{mean}} = \frac{(1 + \beta^2) \cdot P \cdot R}{\beta^2 \cdot P + R} \tag{17}$$

where $F_{\text{mean}}$ is harmonic mean of Precision and Recall, P and R are precision and recall and $\beta$ is a weight parameter that adjusts the importance of precision and recall (Here $\beta = 1$).

### 2.4.6 BERT Score

BERTScore is an evaluation metric designed to assess the quality of generated text by comparing it to reference text using contextualized word embeddings, rather than relying solely on surface-level textual overlaps [61]. Firstly, text is tokenized into smaller units and tokenizers convert raw text into sequence of tokens. The tokenized text is then fed into a pre-trained transformer-based language model, such as BERT (Bidirectional Encoder Representations from Transformers). The BERT model is designed to create contextual embeddings for each token in the input text. These embeddings represent the meaning of the tokens in the context of the entire sentence or passage. Figure 5 illustrates the process of tokenization and generation of embedding for a sentence using BERT model. Unlike traditional n-gram-based metrics (like BLEU and ROUGE), BERTScore leverages pre-trained language models, such as BERT, to compute semantic similarity at the token level. Reference and candidate sentence are tokenized, and embeddings are generated for each of them, which are later used to compute cosine similarity. The BERTScore consists of three primary components i.e. Precision, Recall, and F1-Score and these are based on the cosine similarity between the embeddings of the predicted and candidate sentences.

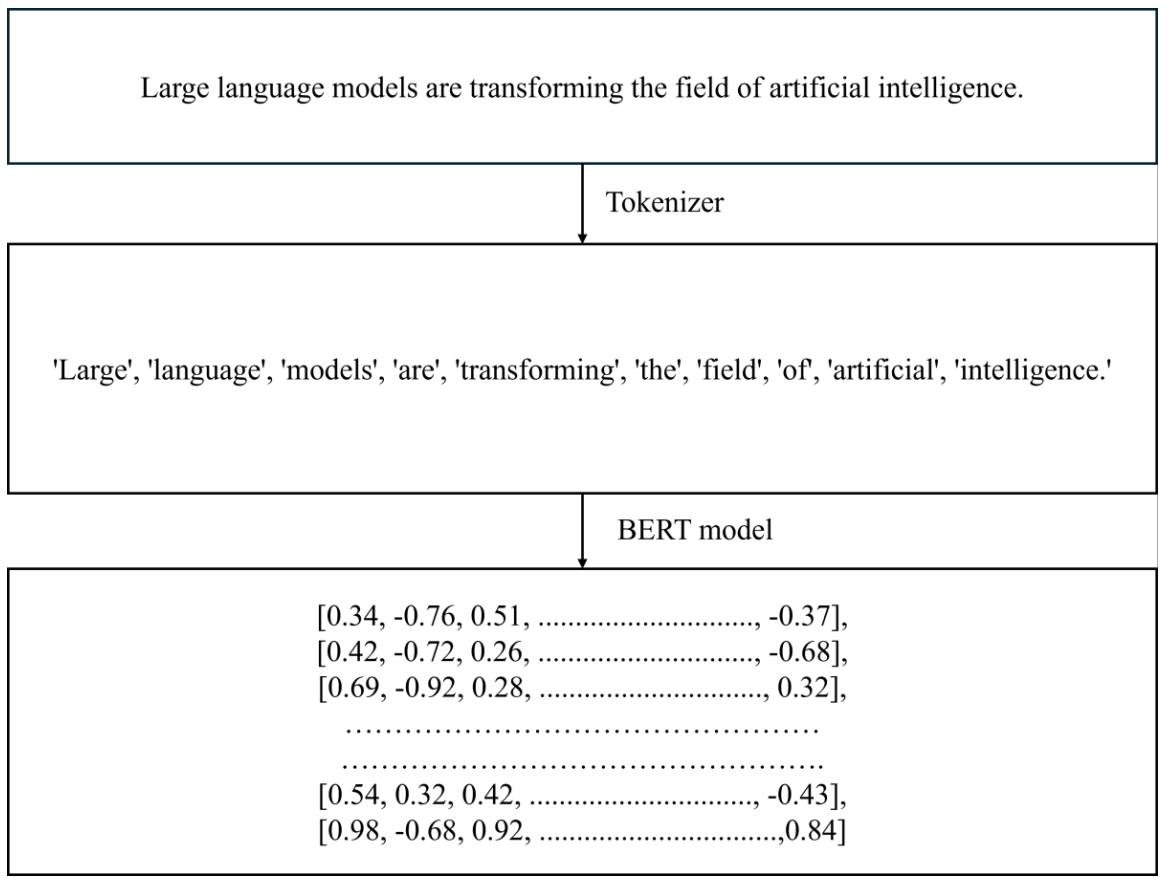

**Figure 5.** The process of tokenization and generation of embeddings.

## 3. Dataset

*3.1 NBCC Introduction*

The NBCC-2020, developed by the Canadian Commission on Building and Fire Codes (CCBFC) and published by the National Research Council of Canada (NRC), establishes the technical requirements for building design and construction across Canada [40]. CCBFC is an independent body set up by NRC and is made of volunteers across the Canada from all facets of Codes-user community. Members of CCBFC include engineers, builders, skilled trade workers, architects, fire and building officials. NBCC provides guidelines for new construction, as well as the alteration, change of use, and demolition of existing structures. It also ensures safety, accessibility, and energy efficiency, serving as a crucial reference for architects, engineers, and regulatory authorities. NBCC is organized into three divisions across two volumes. Division A outlines the Code's scope, objectives, and functional statements, serving as a framework but not a standalone guide for design or compliance. Division B contains the acceptable solutions, which are technical requirements that reflect a minimum acceptable level of performance and are linked to the objectives and functions in Division A. Most users primarily follow Division B. Division C sets out the administrative provisions that govern how the Code is applied and enforced.

*3.2 Dataset Generation*

Dataset for this study is derived from the NBCC, and we have generated 1436 CQA triplets. This dataset was used for testing capabilities of retrievers. Moreover, this data was split for fine-tuning models and testing pre-trained and fine-tuned LLMs. 991 CQA pairs were utilised for fine-tuning pre-trained LLMs while 445 were utilised for testing purpose. By utilising this pre-processed dataset in fine-tuning, we aimed to optimize the text generative capabilities of these LLMs and ensure alignment of LLMs with the specific details and nuances present in NBCC, which in turn can help in optimising the performance of a RAG system. A systematic and concise workflow for generation of dataset is mentioned in Figure 6.

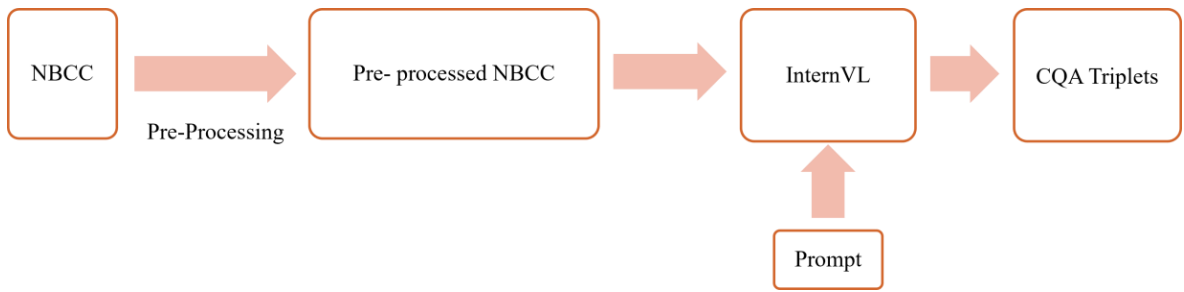

**Figure 6**. Workflow for generation of dataset.

Dataset was meticulously prepared through different steps and model with multi-modality was used to complete the task. First, preprocessed NBCC documents was conducted through the removal of those pages which were not needed for the task. This step ensured that only the most relevant information was to be used for further analysis, hence improving the quality and relevance of the dataset. We use the InternVL2_5-8B to generate the CQA triplets from the pre-processed NBCC documents. InternVL is an open-source multi-modal large language model (MLLM) released by OpenGVLab [62]. The model processes images of pages by first splitting them into smaller, equally sized tiles that preserve spatial content and layout. Each image is resized and divided into a grid of tiles using a dynamic preprocessing strategy based on the image's aspect ratio. These tiles are then individually transformed using normalization and tensor conversion steps to prepare them as input for the model. The preprocessed image tiles are stacked and passed to the InternVL model as visual embeddings alongside the prompt. Prompt instructs the model to generate CQA triplets based on the visual content. The InternVL model, which is capable of multi-modal understanding, uses its vision-language backbone to jointly interpret both the visual features from the image tiles and the semantic intent of the text prompt. As a result, the model can extract relevant textual information from the images and formulate CQA triplets. The specific prompt used is mentioned in Figure 7. Utilised prompt ensured that context was a faithful generation of the text present in pdf and ensured both questions and answers were directly derived from the generated context.

```
prompt = (
"Extract exactly two sets of Context-Question-Answer (CQA) pairs from the provided image. "
"Ensure the context is verbatim extraction of the text present in the image. "
"The answer must be fully contained within the generated context, avoiding any content from tables or additional figures in the image. "
"Use only explicit text from the image to form the CQA pairs. "
"Format the output as follows:\n\n"
"Context: <Generated Context>\n"
"Question: <Generated Question>\n"
"Answer: <Generated Answer>"
)
```

**Figure 7.** Prompt used for generating dataset.

Once CQA triplets are generated, they were converted to a proper JSON format for manipulation and further processing. In the processing phase, we identified problems in the generated triplets. Some triplets were incomplete, lacking information on the Question or Answer fields. The incomplete triplets were removed. Additionally, triplets with less relevant information have been manually filtered to increase the quality and relevance of the dataset. Finally, cleaned and refined dataset of CQA triplets derived out of NBCC was prepared. Through these meticulous preparation and pre-processing of dataset, we ensured that dataset was well-suited for the fine-tuning process. An example of our cleaned and pre-processed dataset that was utilised in fine-tuning is mentioned in the Figure 8.

```
    {
        "Context": "A-3.8.2.5. A-3.8.2.4(1) Access to Storeys Served by Escalators and Moving Walks. In some buildings, escalators and inclined moving walks are installed to provide transportation from one floor level to another floor level so as to increase the capacity to move large numbers of persons. Some buildings located on a sloping site are accessible from street level on more than one storey and an escalator or inclined moving walk is provided for internal movement from floor to floor. In both these situations, people must be provided with an equally convenient means of moving between the same floor levels within the building. This may be accomplished by providing elevators, platform-equipped passenger-elevating devices, or ramps, for example. A-3.8.2.5 Parking Areas. In localities where local regulations or bylaws do not govern the provision of barrier-free parking spaces, the following provides guidance to determine appropriate provisions. If more than 50 parking spaces are provided, parking spaces for use by persons with physical disabilities should be provided in the ratio of one for every 100 parking spaces or part thereof. Parking spaces sloping use by persons with physical disabilities should be provided in the ratio of one for every 100 parking spaces or part thereof. Parking spaces (1) be not less than 2400 mm wide and provided on one side with an access aisle not less than 1 500 mm wide, (2) have a firm, slip-resistant and level surface, (3) be located close to an entrance required to conform to Article 3.8.2.2, (4) be clearly marked as being for the use of persons with physical disabilities, and (5) be identified by a sign located not less than 1 500 mm above ground level, with the International Symbol of Access and the words \u201cPermit Required\u201d (Figure A-3.8.2.5-A).\n\nFigure A-3.8.2.5-A \u201cPermit Required\u201d sign\n\nFigure A-3.8.2.5-B Shared access aisle\n\nAsphalt, concrete and gravel are acceptable parking surfaces. Curb ramps should be not less than 920 mm wide. Parallel parking spaces should be not less than 7 000 mm long.",

        "Question": "What are the requirements for parking spaces for use by persons with physical disabilities in buildings where local regulations or bylbaws do not govern the provision of barrier-free parking spaces?",

        "Answer": "If more than 50 parking spaces are provided, parking spaces for use by persons with physical disabilities should be provided in the ratio of one for every 100 parking spaces or part thereof. Parking spaces should be not less than 2 400 mm wide and provided on one side with an access aisle not less than 1 500 mm wide, have a firm, slip-resistant and level surface, be located close to an entrance required to conform to Article 3.8.2.2, be clearly marked as being for the use of persons with physical disabilities, and be identified by a sign located not less than 1500 mm above ground level, with the International Symbol of Access and the words \u201cPermit Required\u201d."
    },
```

**Figure 8.** A sample from training dataset.

*3.3 Dataset Statistics*

Pre-processed dataset from NBCC was analysed to get a detailed statistical overview of the data. We have generated a plot that depicted the distribution of the lengths of the context in the training dataset as depicted in Figure 9. The frequency on the Y-axis indicates how often certain length ranges appear within the dataset, helping to identify common lengths for each component. The X-axis, representing length in terms of words, shows the range of possible lengths that contexts, questions, and answers can have. The context length distribution reveals the typical amount of information presented in each context, while the question and answer length distributions give an idea of the level of detail expected from the model in both generating and understanding responses. This statistical information is essential for optimizing model performance, ensuring it can efficiently handle both short and long contexts, questions, and answers within the dataset.

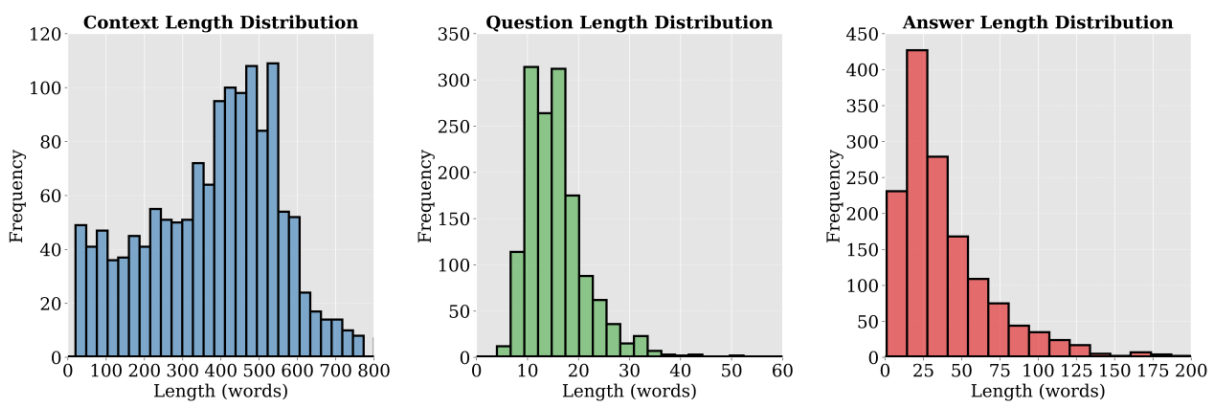

**Figure 9.** Context, question, and answer length distribution

To examine the characteristics of answers in the CQA dataset, answers were categorized as either extractive or abstractive based on their lexical similarity to the corresponding context. In CQA triplet, extractive answers are those that are directly taken from the provided context, while abstractive answers exhibit lower overlap and are more rephrased or synthesized from the context. Simple overlap based measure was employed and answers with more than 50% lexical overlap were categorised as extractive, while those with lower overlap were classified as abstractive. The results of this classification revealed that the dataset contains 1354 extractive answers and 82 abstractive answers. This distribution suggests that the majority of answers in the dataset are extractive in nature, directly drawing from the context, with a smaller proportion being abstractive, where answers are reworded or synthesized. Figure 10 shows the distribution of abstractive and extractive answers.

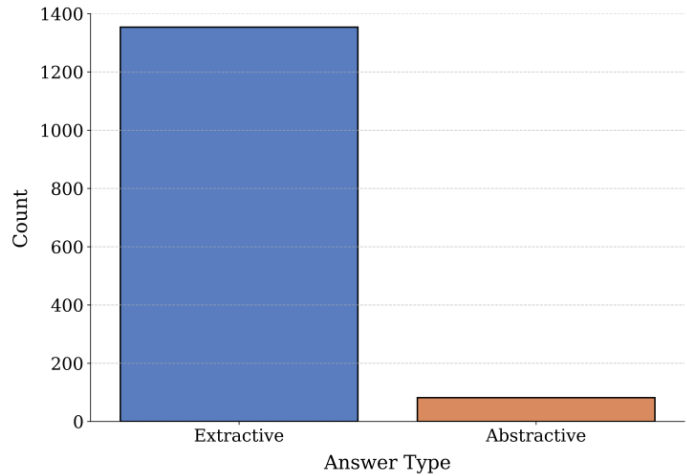

**Figure 10.** Count of extractive and abstractive type of answers.

To further analyze the dataset, trigrams were derived from the questions to identify commonly occurring phrases. Trigrams are sequences of three consecutive words that help capture contextual patterns and commonly used word combinations. This analysis provides insights into the linguistic structure of the questions, highlighting recurring word combinations and potential areas of focus in the dataset. Commonly occurring trigrams can be observed in Figure 11, revealing key linguistic structures that are prominent within the dataset.

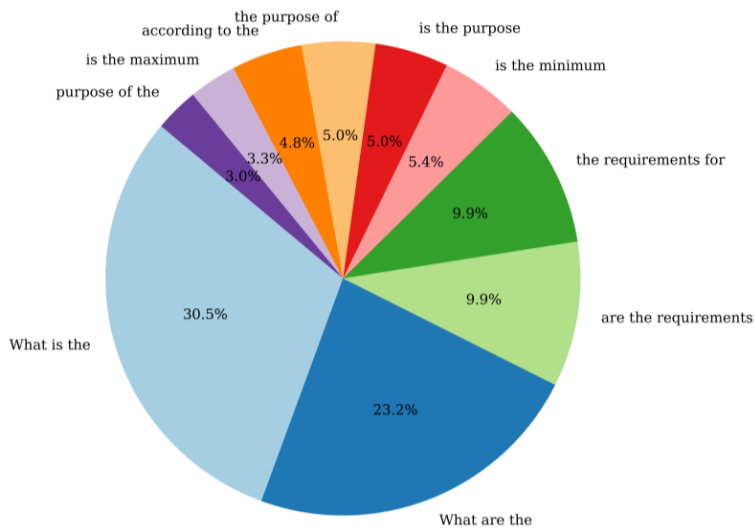

**Figure 11.** Commonly occurring trigrams

# Results and Discussion

## 4.1 Retriever

In this section, we will discuss relative results of the performance of six retriever techniques against their performance on NBCC document using BERT Precision, BERT Recall, and BERT F1-Score for four top-N retrieval settings i.e. top-1, top-3, top-5 and top-10. Table 2 shows performance of different retrieval methods against the BERT based metrics. According to the experiments, ES was the most robust among all retrievers evaluated, performing exceptionally well across all metrics and retrieval settings. The BERT F1 scores for ES were 0.845, 0.849, 0.847, and 0.844 for top-1, top-3, top-5, and top-10 retrieval setting, respectively, which were the highest among all the other retrievers.

**Table 2.** Performance of Various Retrieval Techniques

| Retriever | Average BERT Precision | | | | Average BERT Recall | | | | Average BERT F1-Score | | | |
|---|---|---|---|---|---|---|---|---|---|---|---|---|
| | Top-1 | Top-3 | Top-5 | Top-10 | Top-1 | Top-3 | Top-5 | Top-10 | Top-1 | Top-3 | Top-5 | Top-10 |
| TF-IDF | 0.882 | 0.862 | 0.847 | 0.835 | 0.804 | 0.832 | 0.843 | 0.849 | 0.84 | 0.847 | 0.85 | 0.842 |
| BM-25 | 0.874 | 0.855 | 0.843 | 0.83 | 0.802 | 0.829 | 0.84 | 0.847 | 0.836 | 0.842 | 0.84 | 0.838 |
| ES | 0.884 | 0.862 | 0.848 | 0.836 | 0.81 | 0.837 | 0.847 | 0.852 | 0.845 | 0.849 | 0.85 | 0.844 |
| DPR-FAISS | 0.866 | 0.848 | 0.838 | 0.827 | 0.798 | 0.826 | 0.838 | 0.844 | 0.831 | 0.837 | 0.84 | 0.835 |
| S-BERT-Cos | 0.881 | 0.86 | 0.848 | 0.835 | 0.804 | 0.832 | 0.843 | 0.849 | 0.84 | 0.845 | 0.85 | 0.842 |
| DPR-AI | 0.838 | 0.826 | 0.818 | 0.811 | 0.785 | 0.812 | 0.823 | 0.828 | 0.81 | 0.819 | 0.82 | 0.819 |

### 4.1.1 BERT Precision

BERT Precision measures the proportion of retrieved documents that are relevant to ground truth. A higher precision score indicates that when the retriever retrieves a document, it's more likely to be relevant to the ground truth. It becomes essential when only relevant information is target while keeping irrelevant to be minimum. So, high BERT Precision indicates that retrieved context covers most of the ground truth and irrelevant text are minimum in retrieved context. In terms of BERT precision, ES performed best with a top-1 precision of 0.884. This suggests that ES is highly effective at retrieving relevant documents in the first position. This is likely due to its robust scoring mechanism (BM25) which accounts for term frequency and document length, making it effective at ranking documents based on their relevance.

On the other hand, S-BERT-Cos, TF-IDF and BM25 performed closely to ES, with a top-1 precision of 0.881, 0.882 and 0.874 respectively. S-BERT-Cos uses semantic embeddings and cosine similarity, which allows it to capture the contextual meaning of the query and document. This approach excels in retrieval tasks where the documents have different wording but share a similar context. TF-IDF, while slightly behind ES and S-BERT-Cos in top-1 precision, remains a strong baseline for retrieval tasks. It relies on term frequency and inverse document frequency to assign importance to words, making it effective for keyword-based retrieval but less adept at capturing semantic similarity. BM25, scoring 0.874 in top-1 precision, is a probabilistic ranking function that extends TF-IDF by considering term saturation and document length normalization. Its competitive performance highlights its ability to balance relevance and document ranking, making it a widely used choice for lexical-based retrieval systems.

For DPR-FAISS and DPR-AI, their relatively lower precision scores are likely a result of less effective ranking mechanisms in comparison to ES. Specifically, DPR-AI stands out as the weakest in top-1 precision. Despite its advanced neural architecture, DPR-AI may struggle to prioritize the most relevant documents due to challenges in its ranking mechanism or reliance on embeddings that don't always capture fine-grained relevance well. One key limitation of these methods is the reliance on approximate nearest neighbor (ANN) search, which introduces a degree of inaccuracy. In DPR-FAISS, IndexFlatIP was used and it utilises simple inner product calculations for performing similarity matching. Since IndexFlatIP lacks any clustering or graph-based search, it treats all embeddings equally without prioritizing relevant regions of the vector space. This absence of contextual understanding and reliance solely on numerical similarity can lead to suboptimal retrieval performance. As a result, the retrieved passages may deviate from the true context. Moreover, the poor performance of the DPR-AI is likely due to the simplicity of its indexing method. Annoy relies on approximate nearest neighbor search using trees, and with only 10 trees utilised in our method, it may not effectively capture the complex structure of the embedding space, leading to inaccurate retrieval of relevant passages. Additionally, the method relies on dot product, which doesn't account for more complex semantic relationships between passages and queries, resulting in retrievals that may lack context or relevance.

In summary ES, S-BERT-Cos, TF-IDF, and BM25 excelled across all top-1, top-3, top-5, and top-10 setting. DPR-FAISS and DPR-AI fell behind, largely due to their reliance on approximate search techniques, static embeddings, and less effective ranking mechanisms.

*4.1.2 BERT Recall*

BERT Recall measures a retriever's ability to return all relevant documents, making it especially important for applications that demand high coverage, such as question answering on building codes or legal document retrieval. It measures how much of the ground truth is captured by retrieved context. A high BERT score indicates that most of ground truth is covered by retrieved context along with some irrelevant text. Among the methods evaluated, ES delivers the highest average BERT Recall across all top-N thresholds, with scores of 0.810, 0.837, 0.847, and 0.852 for top-1, top-3, top-5 and top-10 respectively. This consistent performance demonstrates ES's effectiveness in not only retrieving the topmost relevant documents but also capturing a broader set of pertinent results as the retrieval scope expands. ES benefits from its use of the BM25 ranking function, which carefully balances term frequency and document length normalization, making it robust across both exact and partial lexical matches.

Closely following ES are S-BERT-Cos, BM25, and TF-IDF, all of which show comparable recall performance across different top-N setting. For instance, at top-5 and top-10, S-BERT-Cos scores 0.843 and 0.849, BM25 achieves 0.840 and 0.847, and TF-IDF posts 0.843 and 0.849, respectively. Even though underlying mechanism for S-BERT-Cos, BM25 and TF-IDF differs, results have suggested that all three methods are capable of retrieving a substantial proportion of relevant documents. Their similarity in performance underscores the fact that both traditional lexical techniques and modern semantic models have complementary strengths in real-world retrieval tasks such as ours.

On the other hand, DPR-based retrievers, including DPR-FAISS and DPR-AI, lag behind the others in recall performance. DPR-FAISS yields recall scores ranging from 0.798 for top-1 to 0.844 for top-10, while DPR-AI trails slightly further, from 0.785 for top-1 to 0.828 for top-10. Despite their use of dense neural embeddings for semantic matching, these models appear less capable of capturing all relevant documents, particularly when those documents do not align closely in the learned vector space. This highlights a limitation of dense retrieval methods in scenarios where lexical overlap or partial relevance is essential for recall.

In summary, ES remains the top performer in terms of comprehensiveness, but S-BERT-Cos, BM25, and TF-IDF form a strong cluster of high-performing methods, offering balanced trade-offs between semantic and lexical matching. Meanwhile, DPR-FAISS and DPR-AI, while powerful for semantic similarity, may require further tuning or hybridization to match the recall performance of the other approaches.

*4.1.3 BERT F1-Score*

BERT F1-Score provides a balanced measure of retrieval performance by harmonizing precision and recall. It reflects how well a retriever retrieves relevant documents while minimizing irrelevant ones. Among all retrievers, Elasticsearch (ES) consistently achieved the highest F1-Score across top-1, top-3, top-5 and top-10 settings, highlighting its ability to balance relevance and completeness effectively. Its use of the BM25 ranking function enables it to retrieve a wide range of meaningful documents while preserving ranking quality. S-BERT-Cos and TF-IDF ranked a close second, performing strongly in top-3 and top-5 settings, where recall has a greater influence. BM25 showed mid-tier performance, with solid F1-Scores but lacking the semantic depth of S-BERT-Cos and TF-IDF or the scoring refinement of ES.

DPR-FAISS and especially DPR-AI lagged behind. While DPR models can capture semantics, their ranking capabilities were less effective, resulting in lower precision and thus lower F1-Scores. DPR-AI's performance suggests a trade-off where broader recall came at the cost of precision.

In summary, ES emerges as the most balanced and effective retriever across all top-N settings. S-BERT-Cos and TF-IDF is a strong semantic alternative, especially for queries requiring contextual understanding. Other retrievers like BM25 are competent but less refined, while DPR-based methods struggle with maintaining balance between precision and recall. Moreover, the BERT F1-scores suggest that extracting only the top-3 or top-5 relevant documents is sufficient to obtain the desired context, indicating that retrieving more than top-5 documents may not significantly enhance the contextual information and may include irrelevant text.

*4.2 Comparison of pre-trained and fine-tuned models*

The comprehensive performance evaluation of pre-trained and fine-tuned language models on the NBCC dataset reveals significant improvements across multiple evaluation metrics, with nuanced patterns emerging across different model architectures and sizes. Table 3 provides detailed quantitative evidence of these improvements, demonstrating the effectiveness of fine-tuning on specific architectural components. The test methodology evaluates the model's capability to extract and articulate relevant information from retrieved NBCC contexts. The continuous improvement in F1, BLEU, ROUGE, and METEOR scores shows that the model has better understood the complexities of the NBCC content, leading to more accurate, comprehensive, and technically precise responses.

Mistral-Small-24b-Instruct-2501, the largest model evaluated with 24 billion parameters and only 0.39 percent of parameters fine tuned, achieved the highest absolute performance after fine tuning, with an F1 score of 0.688, BLEU of 0.549, and ROUGE 1 of 0.701. This highlights that even minimal fine tuning can lead to improvements, with relative gains of 6.17 percent in F1 score and 11.36 percent in BLEU score. In addition to these fine-tuning results, the pre trained Mistral model also demonstrated exceptional performance. Its BERT F1 score was 0.932, surpassing the fine-tuned BERT F1 scores of all other models. This strong baseline performance of the pre-trained Mistral model can be largely attributed to its massive parameter count, which enables richer representation learning and generalization capabilities. Benchmark results across multiple tasks further support its robustness [63]. For instance, Mistral outperformed GPT 4o mini on the Graduate Level Google Proof Q&A benchmark with chain of thought prompting, scoring 0.453 compared to GPT's 0.377. In the Massive Multitask Language Understanding test with five shot chain of thought prompts, it again surpassed gpt-4o mini by achieving a competitive score of 0.663. Furthermore, in the mtbench dev evaluation, which measures instruction following and response quality, Mistral scored 8.35, slightly higher than GPT 4o mini. These results on benchmarks justify the strong performance of the pre trained Mistral model.

Llama-3.1-8b exhibited exceptional improvement percentages across all key evaluation metrics such as 58.59% in F1 Score, 101.14% in BLEU Score, 53.33% in ROUGE Score-1. It suggests that the targeted adaptation of attention projection matrices using LoRA significantly enhanced its ability to derive accurate answers from provided NBCC context. The dramatic BLEU Score improvement indicates substantially better lexical and phrasal alignment with reference answers, suggesting the model learned to adopt the precise technical language and regulatory phrasing characteristic of NBCC interpretations. Additionally, the BERT F1 score increased by 6.83%, reflecting a notable enhancement in the model's ability to generate answers that are both relevant and accurate compared to the ground truth. The improved evaluation metrics directly correlate with enhanced answer generation capabilities, crucial for automated NBCC question answering systems. The significant F1 Score improvements across all models indicate better precision-recall balance in answer generation, suggesting more accurate and complete responses to building code queries that neither omit critical information nor include irrelevant details. This precision is essential for generating regulatory guidance where both completeness and accuracy are paramount.

Llama-2-7b and Llama-3.2-1b also showed notable improvements after fine tuning, with their F1 scores increasing by 42.80 percent and 42.75 percent respectively. BLEU and ROUGE 1 scores followed a similar trend, with Llama-2-7b improving by 79.53 percent in BLEU and 43.71 percent in ROUGE 1, while Llama-3.2-1b improved by 79.67 percent in BLEU and 41.96 percent in ROUGE 1. Despite these gains post fine-tuning, both models recorded relatively low absolute scores in their pre trained and fine-tuned states. The pre trained F1 scores were 0.271 for Llama-2-7b and 0.269 for Llama-3.2-1b, reflecting limited out of the box performance on NBCC tasks. After fine tuning, the F1 scores increased modestly to 0.387 and 0.384 respectively. BLEU and ROUGE 1 scores also remained relatively low, with 0.228 and 0.411 for fine-tuned Llama-2-7b, and 0.221 and 0.406 for fine-tuned Llama-3.2-1b. Interestingly, although Llama-3.2-1b has significantly fewer parameters, only 1 billion compared to Llama-2-7b's 7 billion, its performance was nearly on par with the larger model across all evaluation metrics. This suggests that Llama-3.2-1b benefits from the architectural advancements introduced in the Llama 3 series. Being released after Llama-2-7b, it likely takes advantage of a pretraining dataset that is seven times larger than that used for Llama 2, along with improved training stability [64]. Furthermore, the Llama 3 series use a tokenizer with a vocabulary of 128K tokens, compared to 32K in Llama 2, resulting in more efficient text encoding and enhanced overall performance.

While fine-tuning with LoRA consistently led to substantial improvements across most models, it is important to recognize that the extent of improvement can vary based on model architecture and parameter size. Notably, Qwen-2.5-7b-Instruct and Phi-3-Mini-4k-Instruct exhibited minor performance declines in specific semantic metrics. Qwen showed a -0.76% drop in BERT Recall, despite improvements in F1, BLEU, and ROUGE scores, suggesting that LoRA-based fine-tuning may have led to task-specific overfitting, limiting the model's ability to maintain semantic diversity in its outputs. Similarly, Phi-3-Mini-4k-Instruct, though achieving a +65.86% improvement in F1 Score and +145.81% in BLEU, showed 3.90 % reduction in BERT Recall and 1.18 % reduction in BERT F1 Score. This could be due to its smaller capacity, making it more susceptible to capacity saturation or imbalance in representational learning during LoRA updates. These observations underscore the importance of architecture-aware fine-tuning strategies, as even efficient adaptation techniques like LoRA require careful calibration to avoid performance regressions in metrics that capture deeper semantic alignment.

Table 3. Performance of pre-trained and fine-tuned models

| Model | | Average F1 Score | Average BLEU Score | Average ROUGE Score-1 | Average SMS | Average METEOR Score | Average BERT Precision | Average BERT Recall | Average BERT F1 Score |
|---|---|---|---|---|---|---|---|---|---|
| Qwen-2.5-7b-Instruct | Pre-Trained | 0.402 | 0.239 | 0.420 | 3.988 | 0.529 | 0.865 | 0.924 | 0.892 |
| | Fine-Tuned | 0.465 | 0.314 | 0.481 | 3.298 | 0.552 | 0.872 | 0.917 | 0.893 |
| | Improvement (%) | 15.67 | 31.38 | 14.52 | 17.30 | 4.35 | 0.81 | -0.76 | 0.11 |
| Llama-2-7b | Pre-Trained | 0.271 | 0.127 | 0.286 | 4.092 | 0.346 | 0.818 | 0.861 | 0.838 |
| | Fine-Tuned | 0.387 | 0.228 | 0.411 | 3.578 | 0.484 | 0.860 | 0.908 | 0.882 |
| | Improvement (%) | 42.80 | 79.53 | 43.71 | 12.56 | 39.88 | 5.13 | 5.46 | 5.25 |
| Llama- 3.1-8b | Pre-Trained | 0.326 | 0.176 | 0.345 | 4.207 | 0.353 | 0.844 | 0.856 | 0.849 |
| | Fine-Tuned | 0.517 | 0.354 | 0.529 | 3.488 | 0.534 | 0.905 | 0.910 | 0.907 |
| | Improvement (%) | 58.59 | 101.14 | 53.33 | 17.09 | 51.28 | 7.23 | 6.31 | 6.83 |
| Llama- 3.2-1b-Instruct | Pre-Trained | 0.269 | 0.123 | 0.286 | 4.067 | 0.347 | 0.823 | 0.873 | 0.846 |
| | Fine-Tuned | 0.384 | 0.221 | 0.406 | 3.639 | 0.479 | 0.860 | 0.907 | 0.882 |
| | Improvement (%) | 42.75 | 79.68 | 41.96 | 10.52 | 38.04 | 4.50 | 3.90 | 4.26 |
| Phi-3-Mini-4k-Instruct | Pre-Trained | 0.290 | 0.131 | 0.303 | 3.903 | 0.352 | 0.828 | 0.871 | 0.848 |
| | Fine-Tuned | 0.481 | 0.322 | 0.497 | 3.872 | 0.477 | 0.840 | 0.837 | 0.838 |
| | Improvement (%) | 65.86 | 145.80 | 64.03 | 0.79 | 35.51 | 1.45 | -3.90 | -1.18 |
| Mistral-Small-24b-Instruct-2501 | Pre-Trained | 0.648 | 0.493 | 0.671 | 2.790 | 0.631 | 0.938 | 0.927 | 0.932 |
| | Fine-Tuned | 0.688 | 0.549 | 0.701 | 2.486 | 0.662 | 0.948 | 0.936 | 0.941 |
| | Improvement (%) | 6.17 | 11.36 | 4.47 | 10.90 | 4.91 | 1.07 | 0.97 | 0.97 |

## 5. Conclusion

This study aimed to evaluate the effectiveness of various retrieval algorithms and the impact of domain-specific fine-tuning on LLMs using the NBCC dataset. Through a structured two-phase experimental setup, we first assessed the retrieval capabilities of different methods in accurately extracting relevant context for queries, followed by an analysis of how fine-tuning pre-trained LLMs using LoRA enhances their generative performance.

In the first set of experiments, ES emerged as the best method for retrieving relevant documents against a query out of the NBCC document, consistently surpassing other techniques across all metrics and retrieval scenarios. It performed better than other retrievers because it utilises the BM25 ranking algorithm for ranking, which is highly efficient for complex documents such as NBCC. Moreover, its inverted index enables keyword matching and phrase matching, which helps in reducing retrieval latency. Additionally, ES supports fine-grained tokenization and normalization, which can handle complex queries more effectively. ES's ability to accurately retrieve the most relevant documents underscores its suitability for tasks requiring high precision and efficient document retrieval, especially crucial when dealing with technical content like the NBCC. When comparing retrieval scenarios for ES, the retrieval of top-3 and top-5 documents performed better than top-1 and top-10 settings across almost all metrics, suggesting that for our dataset, retrieving just the top-3 or top-5 highly relevant documents is sufficient in most cases to address the question.

The second set of experiments analyzed the effect of fine-tuning using the NBCC dataset on several LLMs. Fine-tuning LLMs using LoRA on the NBCC-derived dataset significantly boosted their generative capabilities. The Mistral-Small-24b-Instruct-2501 model emerged as the strongest overall performer. Despite fine-tuning only 0.39% of its parameters, it achieved the highest absolute scores across F1, BLEU, ROUGE metrics and BERT scores. Furthermore, its pre-trained state surpassed the fine-tuned performance of other models, which was obvious as it shows exceptional performance on external benchmarks such as the Google Proof Q&A and MT-Bench. Moreover, Llama-3.1-8b showed the most substantial improvements post fine-tuning, with over 100% gain in BLEU score and more than 50% in F1 and ROUGE-1 scores, underscoring the importance of adapting model internals, such as attention mechanisms, to domain-specific data. While Llama-2-7b and Llama-3.2-1b displayed relatively modest performance due to their lower pre-trained performance, their notable improvements validate the effectiveness of fine-tuning. These results consistently demonstrates that the chosen fine-tuning parameters and configurations customise the LLMs to perform well over NBCC and enhances the model's generational performance across a wide range of evaluation metrics. It is also worth noting that a few models such as Qwen-2.5-7b-Instruct and Phi-3-Mini-4k-Instruct exhibited minor decreases in semantic metrics like BERT Recall and BERT F1 Score. This suggests that while LoRA is highly effective in improving performance, its performance gains may vary depending on model architecture and capacity, highlighting the importance of tuning strategies that balance both syntactic precision and semantic generalization.

Since retriever and LLMs are core component of a RAG framework, therefore it can be concluded that a RAG framework leveraging efficient retrieval methods, such as Elasticsearch, combined with fine-tuned LLMs on the NBCC dataset, can enable the creation of powerful question-answering systems tailored to content of building codes. Such a system would effectively provide quick and intelligent responses to queries out of building codes.

Future works can include the utilization of further optimization techniques and integrations of other advanced models that could really push the boundary on what these systems can achieve. An inclusion of advanced retrievers such as Col-Pali can be included, and their effectiveness can be tested out. Moreover, other features of NBCC, such as images and tables, can be included, and fine-tuning can be done on models with multi-modality. The whole setup of the RAG framework can be built and tested using advanced evaluation metrics, which are meant for RAG.

**Acknowledgement**

This research was funded by the Natural Sciences and Engineering Research Council of Canada (NSERC) through the Alliance grant ALLRP 581074-22.

**Data availability**

Dataset used, fine-tuned models and codes will be made available on request.

# References

[1] M. J. M. Mosenogi, "AN IMPACT ANALYSIS OF CONSTRUCTION SECTOR ON ECONOMIC GROWTH AND HOUSELOLD INCOME IN SOUTH AFRICA".

[2] W. S. Alaloul, M. A. Musarat, M. B. A. Rabbani, Q. Iqbal, A. Maqsoom, and W. Farooq, "Construction Sector Contribution to Economic Stability: Malaysian GDP Distribution," *Sustainability*, vol. 13, no. 9, p. 5012, Apr. 2021, doi: 10.3390/su13095012.

[3] Y. Demir Altıntaş and M. E. Ilal, "Loose coupling of GIS and BIM data models for automated compliance checking against zoning codes," *Autom. Constr.*, vol. 128, p. 103743, Aug. 2021, doi: 10.1016/j.autcon.2021.103743.

[4] J. Wu, X. Xue, and J. Zhang, "Invariant Signature, Logic Reasoning, and Semantic Natural Language Processing (NLP)-Based Automated Building Code Compliance Checking (I-SNACC) Framework," *J. Inf. Technol. Constr.*, vol. 28, pp. 1–18, Jan. 2023, doi: 10.36680/j.itcon.2023.001.

[5] J. Zhang and A. Akanmu, "Intelligent Construction Case Study Illustration System Using Natural Language Processing and Image Searching," 2016.

[6] M. U. Hadi *et al.*, "Large Language Models: A Comprehensive Survey of its Applications, Challenges, Limitations, and Future Prospects," Nov. 16, 2023. doi: 10.36227/techrxiv.23589741.v4.

[7] Y. Liu, J. Cao, C. Liu, K. Ding, and L. Jin, "Datasets for Large Language Models: A Comprehensive Survey," Feb. 28, 2024, *arXiv*: arXiv:2402.18041. doi: 10.48550/arXiv.2402.18041.

[8] K. Denecke, R. May, LLMHealthGroup, and O. Rivera Romero, "Potential of Large Language Models in Health Care: Delphi Study," *J. Med. Internet Res.*, vol. 26, p. e52399, May 2024, doi: 10.2196/52399.

[9] V. Liventsev, A. Grishina, A. Härmä, and L. Moonen, "Fully Autonomous Programming with Large Language Models," in *Proceedings of the Genetic and Evolutionary Computation Conference*, Lisbon Portugal: ACM, Jul. 2023, pp. 1146–1155. doi: 10.1145/3583131.3590481.

[10] A. Louis, G. Van Dijck, and G. Spanakis, "Interpretable Long-Form Legal Question Answering with Retrieval-Augmented Large Language Models," *Proc. AAAI Conf. Artif. Intell.*, vol. 38, no. 20, pp. 22266–22275, Mar. 2024, doi: 10.1609/aaai.v38i20.30232.

[11] K. He *et al.*, "A survey of large language models for healthcare: from data, technology, and applications to accountability and ethics," *Inf. Fusion*, vol. 118, p. 102963, Jun. 2025, doi: 10.1016/j.inffus.2025.102963.

[12] Y. Chang *et al.*, "A Survey on Evaluation of Large Language Models," *ACM Trans. Intell. Syst. Technol.*, vol. 15, no. 3, pp. 1–45, Jun. 2024, doi: 10.1145/3641289.

[13] OpenAI *et al.*, "GPT-4 Technical Report," Mar. 04, 2024, *arXiv*: arXiv:2303.08774. doi: 10.48550/arXiv.2303.08774.

[14] J. Devlin, M.-W. Chang, K. Lee, and K. Toutanova, "BERT: Pre-training of Deep Bidirectional Transformers for Language Understanding," May 24, 2019, *arXiv*: arXiv:1810.04805. doi: 10.48550/arXiv.1810.04805.

[15] A. Chowdhery *et al.*, "PaLM: Scaling Language Modeling with Pathways".

[16] H. Touvron *et al.*, "LLaMA: Open and Efficient Foundation Language Models," Feb. 27, 2023, *arXiv*: arXiv:2302.13971. doi: 10.48550/arXiv.2302.13971.

[17] M. T. R. Laskar, M. S. Bari, M. Rahman, M. A. H. Bhuiyan, S. Joty, and J. X. Huang, "A Systematic Study and Comprehensive Evaluation of ChatGPT on Benchmark Datasets," Jul. 05, 2023, *arXiv*: arXiv:2305.18486. doi: 10.48550/arXiv.2305.18486.

[18] A. Zharovskikh, "Applications of Large Language Models," InData Labs. Accessed: Feb. 12, 2025. [Online]. Available: https://indatalabs.com/blog/large-language-model-apps

[19] M. Adil, G. Lee, V. A. Gonzalez, and Q. Mei, "Using Vision Language Models for Safety Hazard Identification in Construction".

[20] H. Li, R. Yang, S. Xu, Y. Xiao, and H. Zhao, "Intelligent Checking Method for Construction Schemes via Fusion of Knowledge Graph and Large Language Models," *Buildings*, vol. 14, no. 8, p. 2502, Aug. 2024, doi: 10.3390/buildings14082502.

[21] X. Guo, P. K.-Y. Wong, J. C. P. Cheng, J. C. F. Chan, P.-H. Leung, and X. Tao, "Enhancing Visual-Llm Through Prompt Engineering and a Two-Stage Retrieval-Augmented Generation Algorithm for Construction Site Safety Compliance Checking," 2025, *SSRN*. doi: 10.2139/ssrn.5097308.

[22] "ErgoChat – a Visual Query System for the Ergonomic Risk Assessment of Construction Workers."

[23] S. Wong, C. Zheng, X. Su, and Y. Tang, "Construction contract risk identification based on knowledge-augmented language models," *Comput. Ind.*, vol. 157–158, p. 104082, May 2024, doi: 10.1016/j.compind.2024.104082.

[24] G. Chen, A. Alsharef, A. Ovid, A. Albert, and E. Jaselskis, "Meet2Mitigate: An LLM-powered framework for real-time issue identification and mitigation from construction meeting discourse," *Adv. Eng. Inform.*, vol. 64, p. 103068, Mar. 2025, doi: 10.1016/j.aei.2024.103068.

[25] N. Chen, X. Lin, H. Jiang, and Y. An, "Automated Building Information Modeling Compliance Check through a Large Language Model Combined with Deep Learning and Ontology," *Buildings*, vol. 14, no. 7, p. 1983, Jul. 2024, doi: 10.3390/buildings14071983.

[26] T. Yu, A. Xu, and R. Akkiraju, "In Defense of RAG in the Era of Long-Context Language Models," Sep. 03, 2024, *arXiv*: arXiv:2409.01666. doi: 10.48550/arXiv.2409.01666.

[27] F. Yang *et al.*, "Empower Large Language Model to Perform Better on Industrial Domain-Specific Question Answering," Oct. 16, 2023, *arXiv*: arXiv:2305.11541. doi: 10.48550/arXiv.2305.11541.

[28] Z. Xu, S. Jain, and M. Kankanhalli, "Hallucination is Inevitable: An Innate Limitation of Large Language Models," Jan. 22, 2024, *arXiv*: arXiv:2401.11817. doi: 10.48550/arXiv.2401.11817.

[29] K. Shuster, S. Poff, M. Chen, D. Kiela, and J. Weston, "Retrieval Augmentation Reduces Hallucination in Conversation," Apr. 15, 2021, *arXiv*: arXiv:2104.07567. doi: 10.48550/arXiv.2104.07567.

[30] Y. Zhang, M. Khalifa, L. Logeswaran, M. Lee, H. Lee, and L. Wang, "Merging Generated and Retrieved Knowledge for Open-Domain QA," Oct. 22, 2023, *arXiv*: arXiv:2310.14393. doi: 10.48550/arXiv.2310.14393.

[31] Z. Han, C. Gao, J. Liu, J. Zhang, and S. Q. Zhang, "Parameter-Efficient Fine-Tuning for Large Models: A Comprehensive Survey," Sep. 16, 2024, *arXiv*: arXiv:2403.14608. doi: 10.48550/arXiv.2403.14608.

[32] K. Rangan and Y. Yin, "A fine-tuning enhanced RAG system with quantized influence measure as AI judge," *Sci. Rep.*, vol. 14, no. 1, p. 27446, Nov. 2024, doi: 10.1038/s41598-024-79110-x.

[33] N. Ding *et al.*, "Parameter-efficient fine-tuning of large-scale pre-trained language models," *Nat. Mach. Intell.*, vol. 5, no. 3, pp. 220–235, Mar. 2023, doi: 10.1038/s42256-023-00626-4.

[34] V. Lialin, V. Deshpande, X. Yao, and A. Rumshisky, "Scaling Down to Scale Up: A Guide to Parameter-Efficient Fine-Tuning," Nov. 22, 2024, *arXiv*: arXiv:2303.15647. doi: 10.48550/arXiv.2303.15647.

[35] Q. Li, "Parameter Efficient Fine-Tuning on Selective Parameters for Transformer-Based Pre-Trained Models," in *2024 IEEE International Conference on Multimedia and Expo (ICME)*, Niagara Falls, ON, Canada: IEEE, Jul. 2024, pp. 1–6. doi: 10.1109/ICME57554.2024.10688138.

[36] S. Avidan, G. Brostow, M. Cissé, G. M. Farinella, and T. Hassner, Eds., *Computer Vision – ECCV 2022: 17th European Conference, Tel Aviv, Israel, October 23–27, 2022, Proceedings, Part XXXIII*, vol. 13693. in Lecture Notes in Computer Science, vol. 13693. Cham: Springer Nature Switzerland, 2022. doi: 10.1007/978-3-031-19827-4.

[37] E. J. Hu *et al.*, "LoRA: Low-Rank Adaptation of Large Language Models," Oct. 16, 2021, *arXiv*: arXiv:2106.09685. doi: 10.48550/arXiv.2106.09685.

[38] S. Chen *et al.*, "AdaptFormer: Adapting Vision Transformers for Scalable Visual Recognition".

[39] L. Xu, H. Xie, S.-Z. J. Qin, X. Tao, and F. L. Wang, "Parameter-Efficient Fine-Tuning Methods for Pretrained Language Models: A Critical Review and Assessment," Dec. 19, 2023, *arXiv*: arXiv:2312.12148. doi: 10.48550/arXiv.2312.12148.

[40] N. R. C. Canada, "National Building Code of Canada 2020." Accessed: Mar. 09, 2025. [Online]. Available: https://nrc.canada.ca/en/certifications-evaluations-standards/codes-canada/codes-canada-publications/national-building-code-canada-2020

[41] P. Zhao *et al.*, "Retrieval-Augmented Generation for AI-Generated Content: A Survey," Jun. 21, 2024, *arXiv*: arXiv:2402.19473. doi: 10.48550/arXiv.2402.19473.

[42] S. Robertson and H. Zaragoza, "The Probabilistic Relevance Framework: BM25 and Beyond," *Found. Trends® Inf. Retr.*, vol. 3, no. 4, pp. 333–389, 2009, doi: 10.1561/1500000019.

[43] M. T. Mohammed and O. F. Rashid, "Document retrieval using term term frequency inverse sentence frequency weighting scheme," *Indones. J. Electr. Eng. Comput. Sci.*, vol. 31, no. 3, p. 1478, Sep. 2023, doi: 10.11591/ijeecs.v31.i3.pp1478-1485.

[44] V. Karpukhin *et al.*, "Dense Passage Retrieval for Open-Domain Question Answering," Sep. 30, 2020, *arXiv*: arXiv:2004.04906. doi: 10.48550/arXiv.2004.04906.

[45] Y. Bai *et al.*, "SparTerm: Learning Term-based Sparse Representation for Fast Text Retrieval," Oct. 02, 2020, *arXiv*: arXiv:2010.00768. doi: 10.48550/arXiv.2010.00768.

[46] L. D. Krisnawati, A. W. Mahastama, S.-C. Haw, K.-W. Ng, and P. Naveen, "Indonesian-English Textual Similarity Detection Using Universal Sentence Encoder (USE) and Facebook AI Similarity Search (FAISS)," *CommIT Commun. Inf. Technol. J.*, vol. 18, no. 2, pp. 183–195, Sep. 2024, doi: 10.21512/commit.v18i2.11274.

[47] N. Reimers and I. Gurevych, "Sentence-BERT: Sentence Embeddings using Siamese BERT-Networks," in *Proceedings of the 2019 Conference on Empirical Methods in Natural Language Processing and the 9th International Joint Conference on Natural Language Processing (EMNLP-IJCNLP)*, Hong Kong, China: Association for Computational Linguistics, 2019, pp. 3980–3990. doi: 10.18653/v1/D19-1410.

[48] A. Smith, "Enhanced support for citations on GitHub," The GitHub Blog. Accessed: Feb. 20, 2025. [Online]. Available: https://github.blog/news-insights/company-news/enhanced-support-citations-github/

[49] A. Vaswani *et al.*, "Attention is All you Need".

[50] J. Lang, Z. Guo, and S. Huang, "A Comprehensive Study on Quantization Techniques for Large Language Models," in *2024 4th International Conference on Artificial Intelligence, Robotics, and Communication (ICAIRC)*, Xiamen, China: IEEE, Dec. 2024, pp. 224–231. doi: 10.1109/ICAIRC64177.2024.10899941.

[51] "Meta Llama 2," Meta Llama. Accessed: Mar. 14, 2025. [Online]. Available: https://www.llama.com/llama2/

[52] "Introducing Meta Llama 3: The most capable openly available LLM to date," Meta AI. Accessed: Mar. 14, 2025. [Online]. Available: https://ai.meta.com/blog/meta-llama-3/

[53] "Qwen LLMs." Accessed: Mar. 14, 2025. [Online]. Available: https://www.alibabacloud.com/help/en/model-studio/developer-reference/what-is-qwen-llm

[54] "mistralai/Mistral-Small-24B-Instruct-2501 · Hugging Face." Accessed: Mar. 29, 2025. [Online]. Available: https://huggingface.co/mistralai/Mistral-Small-24B-Instruct-2501

[55] M. Bilenko, "Introducing Phi-3: Redefining what's possible with SLMs," Microsoft Azure Blog. Accessed: Mar. 14, 2025. [Online]. Available: https://azure.microsoft.com/en-us/blog/introducing-phi-3-redefining-whats-possible-with-slms/

[56] Z. C. Lipton, C. Elkan, and B. Narayanaswamy, "Thresholding Classifiers to Maximize F1 Score," May 14, 2014, *arXiv*: arXiv:1402.1892. doi: 10.48550/arXiv.1402.1892.

[57] K. Papineni, S. Roukos, T. Ward, and W.-J. Zhu, "BLEU: a method for automatic evaluation of machine translation," in *Proceedings of the 40th Annual Meeting on Association for Computational Linguistics - ACL '02*, Philadelphia, Pennsylvania: Association for Computational Linguistics, 2001, p. 311. doi: 10.3115/1073083.1073135.

[58] C.-Y. Lin, "ROUGE: A Package for Automatic Evaluation of Summaries".

[59] E. Clark, A. Celikyilmaz, and N. A. Smith, "Sentence Mover's Similarity: Automatic Evaluation for Multi-Sentence Texts," in *Proceedings of the 57th Annual Meeting of the Association for Computational Linguistics*, Florence, Italy: Association for Computational Linguistics, 2019, pp. 2748–2760. doi: 10.18653/v1/P19-1264.

[60] S. Banerjee and A. Lavie, "METEOR: An Automatic Metric for MT Evaluation with Improved Correlation with Human Judgments," Jun. 2005, [Online]. Available: https://aclanthology.org/W05-0909/

[61] T. Zhang, V. Kishore, F. Wu, K. Q. Weinberger, and Y. Artzi, "BERTScore: Evaluating Text Generation with BERT," Feb. 24, 2020, *arXiv*: arXiv:1904.09675. doi: 10.48550/arXiv.1904.09675.

[62] "OpenGVLab/InternVL2_5-8B · Hugging Face." Accessed: Apr. 04, 2025. [Online]. Available: https://huggingface.co/OpenGVLab/InternVL2_5-8B

[63] "mistralai/Mistral-Small-24B-Instruct-2501 · Hugging Face." Accessed: Apr. 23, 2025. [Online]. Available: https://huggingface.co/mistralai/Mistral-Small-24B-Instruct-2501

[64] "Introducing Meta Llama 3: The most capable openly available LLM to date," Meta AI. Accessed: Apr. 29, 2025. [Online]. Available: https://ai.meta.com/blog/meta-llama-3/

.